\documentclass{article}

\usepackage{arxiv}

\usepackage[utf8]{inputenc} % allow utf-8 input
\usepackage[T1]{fontenc}    % use 8-bit T1 fonts
\usepackage{hyperref}       % hyperlinks
\usepackage{url}            % simple URL typesetting
\usepackage{booktabs}       % professional-quality tables
\usepackage{amsfonts}       % blackboard math symbols
\usepackage{nicefrac}       % compact symbols for 1/2, etc.
\usepackage{microtype}      % microtypography
\usepackage{lipsum}
\usepackage{fancyhdr}       % header
\usepackage{graphicx}       % graphics
\graphicspath{{media/}}     % organize your images and other figures under media/ folder
\usepackage{algorithm}
\usepackage{algcompatible}
\usepackage{algpseudocode}
\usepackage{amsmath}
\usepackage{hyperref}
\usepackage{float}
\usepackage{enumitem}
\usepackage{tikz}
\usepackage{verbatim}
\usepackage{subcaption}
\usepackage{subfloat}
\usepackage[fontsize=10.8pt]{fontsize}

\newcommand*\circled[1]{%
  \tikz[baseline=(char.base)]{%
    \node[
      shape=circle,
      draw,
      inner sep=0.5pt,
      fill=black,
      text=white
    ] (char) {#1};}}

\newcommand*\circledwhite[1]{%
  \tikz[baseline=(char.base)]{%
    \node[
      shape=circle,
      draw,
      inner sep=0.5pt,
    ] (char) {#1};}}

\normalsize
\newcommand{\uppermark}[1]{$^{#1}$}

\title{
Efficient Mixture-of-Agents Serving via Tree-Structured Routing, Adaptive Pruning, and Dependency-Aware Prefill-Decode Overlap
}
\author{
  \vspace{1pt} Zijun Wang\uppermark{1,}\thanks{The first two authors contributed equally to this paper.} \quad  Yijiahao Qi\uppermark{2,}\footnotemark[1] \quad  Hanqiu Chen\uppermark{1} \quad Zishen Wan\uppermark{1}\\ \vspace{2pt} \textbf{Gongjin Sun\uppermark{3} \quad Dongyang Li\uppermark{3} \quad Shuyi Pei\uppermark{3} \quad Cong  Hao\uppermark{1}}\\
  \vspace{1pt} \uppermark{1}Georgia Institute of Technology, \ \uppermark{2}Peking University, \ \\  \vspace{1pt}\uppermark{3}Samsung Semiconductors, Inc.\\
   \texttt{\{zwang3547, hanqiu.chen, zishenwan, callie.hao\}@gatech.edu} \\
}

\begin{document}
\maketitle

\begin{abstract}
Mixture-of-Agents (MoA) inference can suffer from dense inter-agent communication and low hardware utilization, which jointly inflate serving latency. We present a serving design that targets these bottlenecks through an algorithm-system co-design. First, we replace dense agent interaction graphs with a hierarchical tree topology that induces structured sparsity in inter-agent communication. Second, we introduce a runtime adaptive early exit mechanism that selectively terminates or skips downstream agent invocations using semantic agreement and confidence signals from intermediate outputs. Third, we pipeline agent execution by overlapping incremental prefilling with decoding across dependency-related agents, improving utilization and reducing inference latency. Across representative tasks, this approach substantially reduces end-to-end latency (up to 90\%) while maintaining comparable accuracy (within $\pm$1\%) relative to dense-connectivity MoA baselines, and can improve accuracy in certain settings.
\end{abstract}

% keywords can be removed
% \keywords{First keyword \and Second keyword \and More}
%\vspace{-3mm}
\section{Introduction}
\label{sec:intro}

Large language models (LLMs) are widely used across diverse natural language processing (NLP) tasks. A growing trend in agentic AI is to combine multiple LLMs with complementary strengths as different agents to enhance reasoning~\cite{li2024survey}. Among such LLM-based multi-agent paradigms, Mixture-of-Agents (MoA) is a powerful style that uses multiple LLMs (agents) as parallel proposers, each generating its own answer, and then uses an aggregator model to fuse these outputs into a final response. MoA has demonstrated strong empirical gains in reasoning, QA, and coding~\cite{Jiang2023LLMBlender,Chen2024Reconcile,wan2025reca}.
Although prior works~\cite{Du2023MAD,Shen2023HuggingGPT,Park2023GenerativeAgents,Liu2024DLAN,wang2025slm} have demonstrated the promise and effectiveness of MoA-style multi-agent systems across tasks such as planning, problem solving, debate, critique, and adaptive team formation, two major challenges remain for improving MoA system performance at both the software and hardware levels.

\begin{figure}[t]
  \centering
    \begin{subfigure}[b]{0.48\linewidth}
      \centering
      \includegraphics[width=\linewidth]{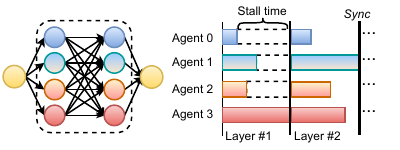}
      \caption{}
      \label{fig:all-to-all}
    \end{subfigure}
  \hfill
    \begin{subfigure}[b]{0.48\linewidth}
      \centering
      \includegraphics[width=\linewidth]{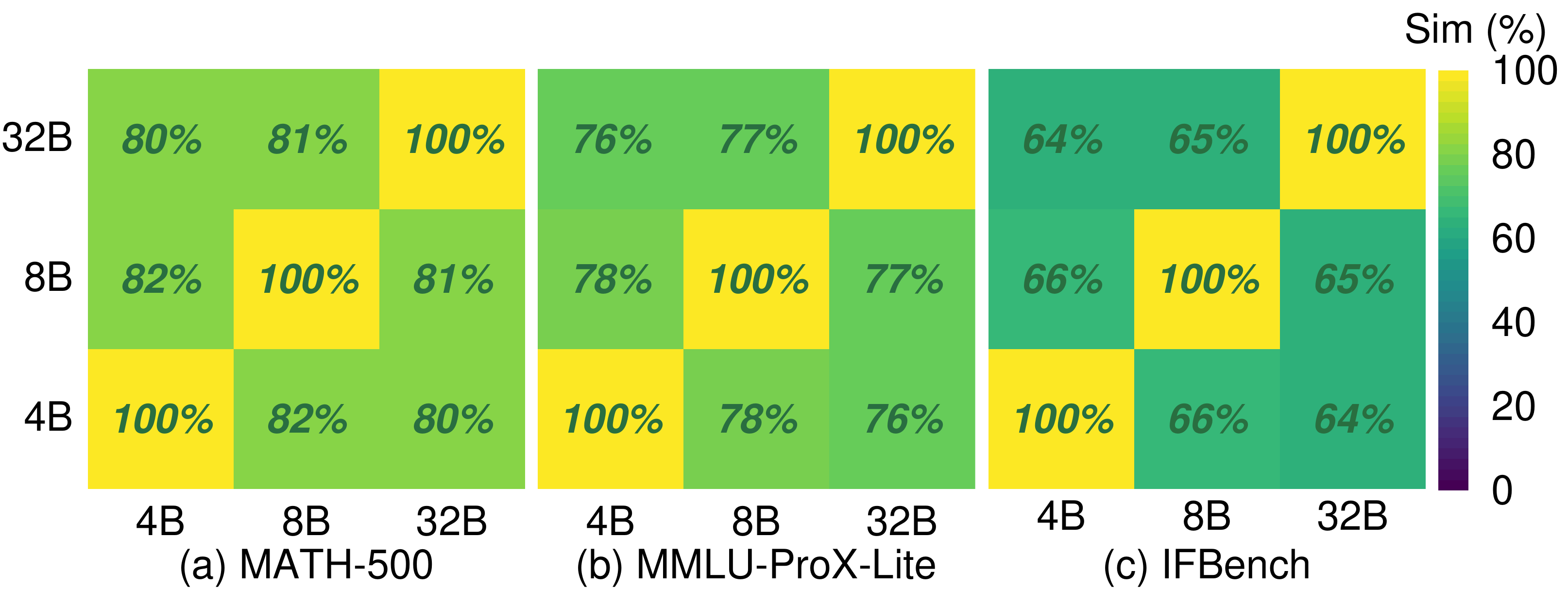}
      \caption{}
      \label{fig:sim_heatmap}
    \end{subfigure}
  \hfill
    \begin{subfigure}[b]{0.48\linewidth}
      \centering
      \includegraphics[width=\linewidth]{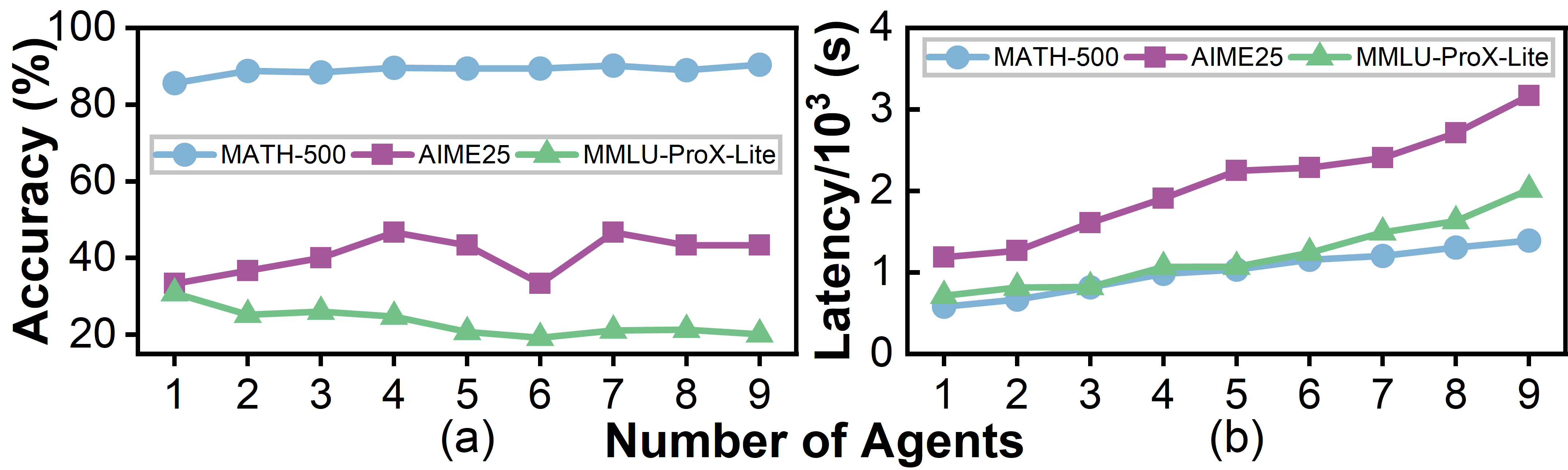}
      \caption{}
      \label{fig:width_saturation}
    \end{subfigure}
  \caption{(1) Conventional all-to-all connected MoA system. Layer synchronization for all agents leads to pipeline stall time, causing low inference efficiency. (2) Semantic similarity within clusters of 1st layer of the tree topology MoA. (3) The relationship between the number of agents, task accuracy and end-to-end latency using all-to-all connection.}
  \label{fig:all_three}
\end{figure}

\circled{1} \textbf{Redundant connections among agents.} Most existing MoA systems adopt an \textbf{all-to-all agent connection topology}, as illustrated in Fig.~\ref{fig:all_three}(\subref*{fig:all-to-all}). These  MoA systems are typically organized into multiple layers, each containing several agents. Agents in adjacent layers are fully connected, with each agent receiving the outputs of all agents in the preceding layer. Although this connection topology intuitively preserves all information exchanged among agents, many of these connections do not transmit meaningful messages, thus may have limited contributions to the quality of the final output. As a result, intelligently orchestrating the connectivity structure and dynamically pruning unnecessary agent connections can not only substantially reduce redundant inter-agent communication for response time improvement, but also preserve the quality of intermediate results so that downstream agents will not be overwhelmed by unnecessary, low-quality, or misleading information.

\circled{2} \textbf{Inefficient MoA system serving on hardware}. 
The hardware inefficiency arises from two sources. First, the all-to-all agent connectivity widely used in existing MoA systems introduces substantial GPU poor computation resource utilization. As shown in Fig.~\ref{fig:all_three}(\subref*{fig:all-to-all}), this is because each agent consumes outputs from all agents in the previous layer, and one layer’s latency is determined by the slowest agent request. Second, existing LLM serving frameworks with prefill-decode (PD) disaggregation provide limited support for efficient MoA execution. They overlook both the heterogeneous latencies of different agents and their complex inter-agent data dependencies, treating the precursor agent’s decoding and the successor agent’s prefilling as strictly sequential operations rather than exploiting potential overlapping opportunities.

Motivated by these two challenges, in this paper, we present \textbf{Faster-MoA}, a unified \emph{algorithm-system} co-design for efficient and low-latency MoA system serving. Our major contributions can be summarized as follows.

\begin{enumerate}[leftmargin=5mm, label=\protect\circled{\arabic*}]
  %\vspace{-1mm}
  \item \textbf{Topology innovation --- Hierarchical tree structure for agents connection.} Instead of the conventional all-to-all connectivity in MoA systems, we propose a novel hierarchical tree topology with two key features. Agents within each layer are grouped into clusters, and each agent in the next layer connects only to a specific cluster for localized information aggregation. The number of agents in the next layer matches the number of clusters in the previous layer. This structure enables efficient local aggregation followed by global aggregation, producing high-quality outputs while avoiding the redundant information exchange in fully connected designs.
  
   \item \textbf{Algorithm innovation --- Semantic-guided run-time dynamic early-exit.} Different tasks exhibit varying levels of difficulty, requiring different numbers and sizes of LLM agents. To dynamically prune unnecessary inter-agent connections at run time based on output similarity and confidence, we introduce a novel agent-level early-exit (EE) mechanism. This mechanism evaluates the semantic similarity and confidence of intermediate outputs from small-LLM agents on the fly. When these small-size agents produce sufficiently high-quality outputs, we prune the connections to the large-LLM agents and proceed to the next layer’s inference without waiting for them to finish. This strategy can largely reduce the GPU idle time for waiting. 
   
  \item \textbf{System innovation --- Agent dependency-aware incremental prefilling for PD overlapping.} To further improve hardware efficiency, we overlap decoding and prefilling stages of agents with data dependencies. We divide precursor agent’s decoding phase into smaller sub-stages and, after each sub-stage, immediately stream the generated tokens to the successor agent for prefilling. This increases the overlap between decoding and prefilling computations for higher GPU resource utilization. 

  \item \textbf{Evaluation.} We evaluate the performance of Faster-MoA on five popular benchmarks with model deployment on NVIDIA H200 GPUs. In comparison with baseline all-to-all connected MoA systems, Faster-MoA achieves 73\% to 90\% decrease in end-to-end (E2E) latency with similar (only $\pm$1\% variation) or even higher task accuracy.
 
\end{enumerate}

\section{Preliminary and Related Works}
\subsection{Topology of MoA Systems}
MoA instantiates an ensemble of proposer agents whose generations are fused by an aggregator agent, sometimes over multiple layers. In such agent organization, the definition of agent dependency between proposer and aggregator agents is defined as follows. Let $o_1, o_2, o_3$ denote the generated texts from three distinct proposer agents $A_1, A_2, A_3$, all connecting to an aggregator agent $A_{\text{agg}}$, then the input prompt $S(A_\text{agg})$ of the aggregator agent is assembled by:
\[
S(A_{\text{agg}}) = \cup (S_{\text{prefix}}, \ o_1, \ o_2, \ o_3, S_{\text{suffix}})
\]
where $S_{\text{prefix}}$ and $S_{\text{suffix}}$ are the prefix and suffix prompts. This "output fusion" behavior is the core of agent dependency, and also the root cause of many workflow inefficiencies. As we mentioned in Sec.~\ref{sec:intro}, many prevailing MoA debating systems \cite{Du2023MAD, liang-etal-2024-encouraging, Xiong_2023_debate} adopted \textbf{an all-to-all topology}, which means each aggregator agent in the subsequent layer is connected to \emph{all} agents in the previous layer, just like fully-connected layer in neural networks. However, the effectiveness and contribution of one agent to its successor agent remains unexplored, and the E2E latency of such system is very high \cite{liang-etal-2024-encouraging}. Besides, as presented in Fig.~\ref{fig:all_three}(\subref*{fig:all-to-all}), the slowest agent request always determines the latency of one layer, slowing down the entire system with low GPU utilization because of the stall time. Therefore, dynamically pruning agent dependencies to keep and transfer only important information based on intermediate outcomes becomes an essential and vital topic. Although recent works~\cite{Liu2024DLAN, zhang2024cutcrapeconomicalcommunication, chen2023agentversefacilitatingmultiagentcollaboration} have shown great potential in on-the-fly LLM active agent selection and collaborative agent communication pruning, they still ignore the hardware and system efficiency by only focusing on task accuracy. Drawn by these insights, we introduce a \textbf{hierarchical tree-structured agent topology} (Sec.~\ref{sec:tree}) and a \textbf{semantic-informed, run-time dynamic agent-level early-exit (EE) mechanism} (Sec.~\ref{sec:ee}), both of which significantly improve hardware efficiency while maintaining task accuracy comparable to all-to-all connectivity.

\subsection{PD Disaggregation in LLM Serving}

In modern LLM serving infrastructures, a single user inference request is typically decomposed into a \emph{prefill} stage and a \emph{decode} stage, which are often mapped to distinct sets of GPUs and executed asynchronously. Recent works \cite{Patel2024SplitWise, Zhong2024DistServe} have demonstrated that this PD disaggregation yields both high throughput and substantial flexibility in capacity planning. In particular, the two-stage decomposition naturally enables task-level pipelining: once a request completes its prefill stage and enters decode stage, the prefilling of the next request can immediately begin on the prefill engine, thereby overlapping computation and improving hardware utilization. However, this pipeline implicitly assumes that \emph{each request’s input is independent of all others}, i.e., one request’s prompt does not depend on another request’s generated output. This assumption fails in MoA systems, where in most cases agent requests need to follow certain dependencies, causing one request cannot start to prefill until all dependent results from previous agents are collected. As a result, naive PD disaggregated MoA serving is forced into a largely sequential schedule, where a dependent agent’s prefill must wait until all necessary upstream results are fully materialized, severely limiting potential overlap. 

To our best knowledge, existing LLM serving frameworks do not explicitly address agent dependency-aware orchestration under PD disaggregation in MoA settings. Motivated by this gap, we propose \textbf{an agent dependency-aware prefill–decode overlapping mechanism} (Sec.~\ref{sec:overlap}) that exploits the key-value (KV) handoff boundary between stages to partially hide the prefilling cost of dependent agents behind their precursors’ decoding.

\section{Feasibility Analysis}
\subsection{Explorations on Tree Topology}
The vanilla all-to-all connected MoA topology tends to accumulate redundant rationales and suffer from high E2E latency without proportional gains in accuracy. Tree-structured architectures have a long history in modular and mixture-of-experts (MoE) models, and have provisioned efficient solutions in solving non-linear supervised learning tasks \cite{jordan1994hme, bishop2012bayesianhierarchicalmixturesexperts, jiang1999hierarchical}. In LLM era, tree-based search has also emerged as a strong alternative to naive Chain-of-Thoughts (CoT) \cite{wei2022chain}. Tree-of-Thoughts (ToT) explicitly explores a branching search tree of partial solutions and uses lookahead and backtracking to select promising branches, significantly boosting success rates on tasks like Game of 24 and planning puzzles compared to single-chain CoT \cite{yao2023tree}. Graph-of-Thoughts (GoT) further generalizes to arbitrary graphs of “thoughts,” showing that structured, sparse connectivity among intermediate solutions can outperform naive repetition while reducing token usage~\cite{besta2024got}.  

Motivated by the success of tree topology in previous works, we are the first to adopt a hierarchical tree as MoA topology. This is because tree topology is theoretically grounded by a lot of existing works on hierarchical MoEs, empirically supported by tree-/graph-based reasoning frameworks, and crucially matches the needs of latency-sensitive LLM serving where sparse and localized coordination is more beneficial than unconstrained connectivity.

\subsection{Optimizations on Connection Pruning}
\label{sec:pruning}
With the tree topology fixed, the influence of each agent on overall system behavior remains unclear, particularly whether adding more agents and connections actually improves MoA performance. We study this from two perspectives: (i) \textbf{Mutual semantic similarity across agents.} Motivated by prior observations that different agents often produce highly similar outputs~\cite{smith2025comprehensiveanalysislargelanguage, an2025rethinkingsemanticparsinglarge, pandey2025coremeasuringmultiagentllm}, we use Frobenius Cosine Similarity (FCS)~\cite{PRXLife.3.023005,sharma2025on,wang2025selectingauxiliarydataneural} on last-layer hidden states of a shared embedding model (\emph{Qwen3-Embedding-4B}) to measure semantic similarity between each pair of models (Fig.~\ref{fig:all_three}(\subref*{fig:sim_heatmap})). Within all the clusters of 1st layer of tree topology, we observe higher similarity among the 4B, 8B and 32B \emph{Qwen3-VL-Instruct} models on simpler tasks (\emph{MATH-500} \cite{lightman2023lets}, Fig.~\ref{fig:all_three}(\subref*{fig:sim_heatmap}(a))) and lower similarity on harder tasks (\emph{IFBench} \cite{peng2025agentic}, Fig.~\ref{fig:all_three}(\subref*{fig:sim_heatmap}(c))), suggesting that on straightforward tasks, incorporating larger models often adds limited auxiliary information beyond what smaller models already provide. (ii) \textbf{Impact of saturated connections.} We evaluate a one-layer all-to-all MoA with varying width (number of proposer agents) while using an identical \emph{Qwen3-VL-Instruct-4B} as underlying LLM. As shown in Fig.~\ref{fig:all_three}(\subref*{fig:width_saturation}), task accuracy on \emph{MATH-500}, \emph{AIME25} \cite{aime25} and \emph{MMLU-ProX-Lite} \cite{xuan2025mmluprox} quickly saturates or even declines as the number of agents grows, while latency monotonically increases.

These results yield two key takeaways: (1) For relatively simple tasks, aggregating outputs from smaller models is typically sufficient, whereas for more challenging benchmarks it can be beneficial to selectively wait for outputs from larger, stronger models; (2) Simply adding more fully connected agents does not guarantee higher-quality outputs. Consequently, adaptively truncating or dropping straggler agents can preserve competitive performance while largely reducing end-to-end latency.

\subsection{Intuition of Incremental Prefilling}
\label{sec:overlap-motive}
When serving a request, depending on model size and prompt length, the prefill stage can account for up to $10\%$–$30\%$ of E2E latency~\cite{Zhong2024DistServe}. And as discussed, in dependency-coupled systems, naive scheduling launches the successor agent's prefill only \emph{after} all precursor agents finish decoding and their outputs are collected---successor agents idle while waiting, paying the full cost on the critical path. Therefore, based on the tree topology, we further develop a hardware efficient incremental prefilling technique, successfully overlapping the precursor agents decoding and successor agents prefilling, reducing exposed prefilling latency without affecting final outputs (Fig.~\ref{fig:pd-detail}).

The key intuition behind our incremental prefilling is the reusage of KV blocks in the prefilling stage. As illustrated in an example in Fig.~\ref{fig:pd-detail}, agent 3 depends on the outputs of agents 1 and 2. We partition agent 3’s input prompt into three segments: its own prefix, followed by the answer slots for agents 1 and 2. The prefix segment can be prefilled immediately since it has no data dependencies. Once agent 3 receives the first decoded tokens from agent 1, the corresponding answer segment is appended directly after the prefix. Since this new content is contiguous with the already-prefilled prefix, we can reuse the existing KV blocks and compute KV only for the newly appended tokens, significantly reducing computation. Consequently, decoded tokens from agent 1 can be streamed to agent 3 on-the-fly, enabling overlap between decoding (agent 1) and prefilling (agent 3) despite dependency constraints. Note that agent 2 cannot begin incremental prefilling as soon as its first token arrives, because its answer segment is not contiguous with agent 3’s prefix. This autoregressive characteristic requires that agent 2 must wait until agent 1 finishes decoding so that the preceding segment is complete.

\begin{figure*}[t]
    \centering
    \includegraphics[width=\linewidth]{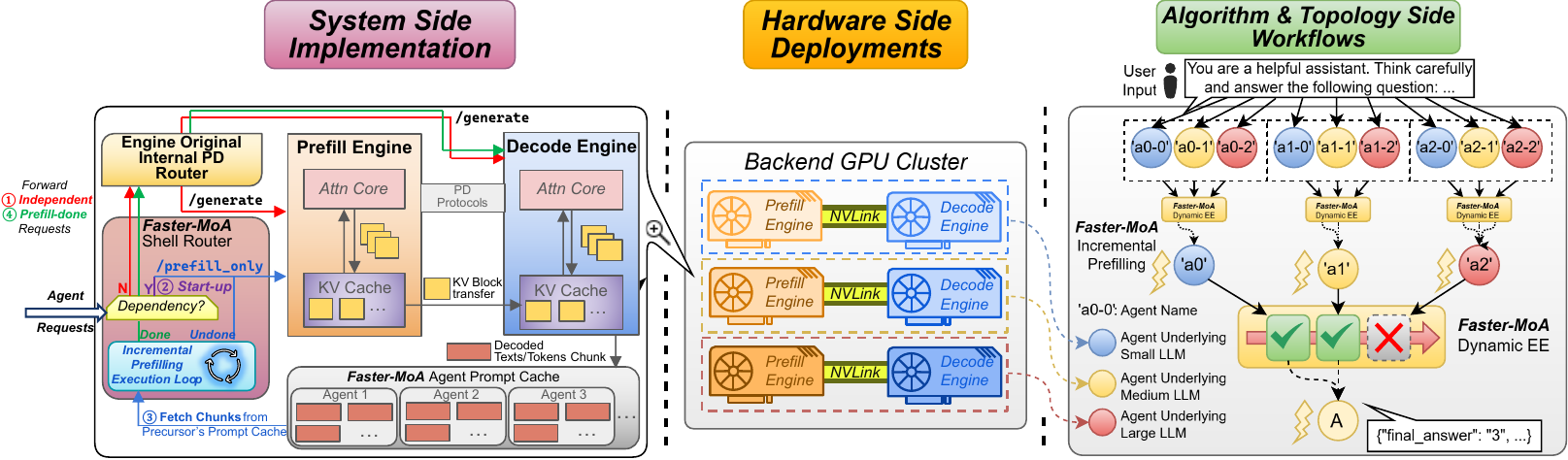}
    %\vspace{-6mm}
    \caption{Faster-MoA overview from three dimensions. We show an example design with three different sizes underlying LLMs. (1) \emph{Algorithm \& topology workflow.} User's input is distributed to three agent clusters with different sizes underlying LLMs. Between layers, we perform agent dependency-aware incremental prefilling for PD overlapping. (2) \emph{Hardware deployments.} Faster-MoA is served upon a 6-GPU cluster. (3) \emph{System Implementation.} Within each pair of PD engines, a dedicated shell router and agent prompt cache are designed for run-time task orchestrations.}
    %\vspace{-4mm}
    \label{fig:1_overview}
    % %\vspace{-5mm}
\end{figure*}

\section{Design Details}
\subsection{Hierarchical Tree Topology}
\label{sec:tree}

As an optimization to all-to-all connections, we propose the \textbf{hierarchical, tree-structured MoA topology}. Fig. \ref{fig:1_overview} (right) illustrates the overall structure and workflow of the framework. 

\textbf{General Definitions.} Let the tree have depth $L$ with leaves at layer $1$ and the root at layer $L$. Input flows from leaves to the root as output. 
Denote the set of agents at layer $\ell$ by $\mathcal{A}_\ell$ and a particular agent by $a_{\ell,j}\in\mathcal{A}_\ell$. 
Each $a_{\ell,j}$ depends only on a \emph{local} subset cluster of precursor agents $\mathcal{C}(a_{\ell,j})\subseteq\mathcal{A}_{\ell-1}$ with total length $|\mathcal{C}(a_{\ell,j})| \ll |\mathcal{A}_{\ell-1}|$. 
This sparsifies dependencies and replaces all-to-all connected layer with \emph{localized readiness}: an agent at layer $\ell$ may proceed as soon as its designated precursors in $\mathcal{C}(a_{\ell,j})$ have produced outputs, independent of unrelated subtrees.

\textbf{Latency implications.} Under an all-to-all design, progress at layer $\ell$ is gated by the slowest agent, yielding a layer inference latency:
%\vspace{-1mm}
\[
T_\ell^{\mathrm{all}} \;=\; \max_{i\in[|\mathcal{A}_\ell|]} t_{\ell,i}
\]
%\vspace{-1mm}
where $t_{\ell,i}$ is the runtime of agent $a_{\ell,i}$. 
The E2E latency accumulates these worst-case layer times, $\sum_{\ell=1}^L T_\ell^{\mathrm{all}}$. 
In the proposed tree topology, each successor agent waits only for its connected precursor agents, so the inference latency of a layer is optimized to: 
%\vspace{-1mm}
\[
T_\ell^{\mathrm{tree}} \;\approx\; \max_{a_{\ell,j}\in\mathcal{A}_\ell}\ \max_{c\in\mathcal{C}(a_{\ell,j})} t_{c},
\]
which is typically much smaller when $|\mathcal{C}(a_{\ell,j})| \ll |\mathcal{A}_{\ell-1}|$.
Besides, the request on different tree branches can also run concurrently, reducing the end-to-end accumulated inference latency from leaf to root, rather than waiting for layer-wise synchronization.

\textbf{Context-length and prefill savings.} Because each precursor agent consumes only $|\mathcal{C}(a_{\ell,j})|$ messages instead of the entire previous layer’s outputs, the input context per precursor shrinks from $|\mathcal{A}_{\ell-1}|$ to $|\mathcal{C}(a_{\ell,j})|$. 
Given that prefill cost scales approximately linearly with prompt length, this yields a proportional reduction in prefill latency and memory traffic. 

\textbf{Redundancy control and variance isolation.} The tree structure naturally reduces redundancy: a successor agent aggregates only its precursors’ outputs rather than the full layer. 
Stragglers or unusually long generations are confined to their subtrees. A slow leaf agent will not delay unrelated branches inference except for its successors, largely improving robustness in long-tailed decoding situations.

As an empirical demonstration, we set 3 precursor agents per successor agent in the tree setting, forming a 9-3-1 three-layered tree structure, with three heterogeneous sized LLMs deployed as underlying model backbones. In this paper, all terminologies related to "tree topology/structure" refer to this setting.

\begin{algorithm}[t!]
    \small
    \caption{Semantic Similarity \& Confidence-based Early Exit}
    \label{alg:MetricQ}
    \begin{algorithmic}[1]
        \Statex Let
        $\langle U,V\rangle_F = \mathrm{trace}(U^\top V)$,
        $\|U\|_F = \sqrt{\langle U,U\rangle_F}$
        \Statex $\mathrm{FrobCosSim}(U,V)
        = \dfrac{\langle \mathrm{Corr}(U), \mathrm{Corr}(V)\rangle_F}
                {\|\mathrm{Corr}(U)\|_F \, \|\mathrm{Corr}(V)\|_F}$

        \Function{MetricQ}{$\{O_i\}_{i=1}^\ell, \{{\log p_\ell^i\}}_{i=1}^{n_a}$}
            \State $C_\ell \gets \exp\!\big(\tfrac{1}{n_a}\sum_{i=1}^{n_a}\log p_\ell^i\big)$ \Comment{Geometric mean confidence}
            % \State $\mathrm{PrevC}.\text{append}(C_\ell)$,\quad $\mathrm{PrevEmb}.\text{append}(T_\ell)$
            \State $\bar{C} \gets \sqrt{\tfrac{1}{\ell}\sum_{i=1}^{\ell} C_i^2} \in [0,1]$ \Comment{RMS average}
            \For{$i = 1$ \textbf{to} $\ell$}
                \State $T_i \gets Embed(O_i), T_\ell \gets Embed(O_\ell)$
                \State $U \gets {T_i}^t \times T_i \in \mathbb{R}^{h \times h} $, $V \gets {T_{\ell}}^t \times T_{\ell} \in \mathbb{R}^{h \times h} $
                \State $\mathrm{Sim}[i,\ell] \gets \mathrm{FrobCosSim}(U, V)$
                \State $\mathrm{Sim}[\ell,i] \gets \mathrm{Sim}[i,\ell]$
            \EndFor
            \State $W \gets \sum_{i=1}^{\ell}\sum_{j=1}^{i} C_i C_j$
            \State $P \gets \frac{1}{W}\sum_{i=1}^{\ell}\sum_{j=1}^{i} C_i C_j \,\mathrm{Sim}[i,j]$
            \State $B \gets 1 - \dfrac{|P - \tau|}{\tau} \in [0,1]$
            \State $Q \gets \sqrt{\bar{C} \cdot B}$
            \State \textbf{Early exit} with probability $Q$ upon receiving $T_\ell$ and $\{\log p_i\}_{i=1}^{n_a}$
        \EndFunction
    \end{algorithmic}
\end{algorithm}

\subsection{Dynamic Early-Exit Routing}
\label{sec:ee}
Algorithm~\ref{alg:MetricQ} shows our innovative agent early-exit mechanism that jointly considers output confidence and semantic-level similarity. The goal is to compute the early-exit probability $Q$ on-the-fly using the outputs of the LLMs that have already completed within the same tree layer. With the computed $Q$, we terminate the remaining unfinished LLMs in the layer and discard their results following the probability $Q$. 

Our algorithm first computes the geometric mean confidence $C_{\ell}$ ($\ell$ stands for the number of currently completed LLMs in a tree layer) from the token-level log-probabilities $\{\log p_\ell^i\}_{i=1}^{n_a}$, which is predicted by agent model, 4B, 8B, and 32B \emph{Qwen3-VL-Instruct} (Line 2), and aggregates all previous confidences into $\bar{C}$ by an RMS average (Line 3), where consistently high-confidence steps yield larger overall confidence $\bar{C}$.

To quantify the semantic similarity of answers that might have different lengths, we first encode each output sentence with a shared embedding model (\emph{Qwen3-Embedding-4B}) (Line 5) and project the last-layer hidden states ($T_i, T_{\ell} \in \mathbb{R}^{n \times h}$) from embedding model into corresponding feature-wise correlation matrix by left multiplying it with its associated transpose matrix (${T_i}^t \times T_i, {T_\ell}^t \times T_\ell \in \mathbb{R}^{h \times h}$) (Line 6). And then, we compute FCS based on these feature-domain matrices (Line 7). All embeddings are produced by a single, shared embedding model so that this similarity is comparable across backbone models from different families. Normalizing matrices into correlation matrices within FCS removes the influence of scales and units, ensuring that they are solely subject to intrinsic semantic variation.

We obtain a universal confidence-weighted similarity score \(P\) by averaging \(\mathrm{Sim}[i,j]\) with corresponding confidence weights \(C_i C_j\) (Line 11), so that steps which are both semantically consistent and high-confidence have the greatest influence, while low-confidence or noisy outputs contribute minimally to entire answer quality.
As mentioned in Sec.~\ref{sec:pruning}, higher similarity is not always better~\cite{an2025rethinkingsemanticparsinglarge,pandey2025coremeasuringmultiagentllm}, we calibrate \(P\) by comparing it against the preference similarity threshold \(\tau\), generating adjusted similarity measure \(B\) (Line 12). The score of $B$ is supposed to favor the scenario when agents are confident and similar enough yet not overly identical, so that their combined responses still offer diverse, complementary information for the next layer. Empirically, we set $\tau$ to 0.7 by default based on our experiment results, thus achieving higher task accuracy with lower agents connectivity compared with other settings.

Finally, Algorithm~\ref{alg:MetricQ} combines this calibrated similarity $B$ with the geometric-mean confidence score $\bar{C}$ to produce the synthesized quality score $Q = \sqrt{\bar{C} \cdot B}$ (Line 13), which stochastically triggers early exit.

\subsection{Incremental Prefilling for PD Overlapping}
\label{sec:overlap}

Fig.~\ref{fig:1_overview} (left \& middle) illustrates our PD disaggregation deployments on GPUs with incremental prefilling. To serve one model, we instantiate two engines running on separate GPUs within the same node. Each engine maintains its own attention compute core and an engine-local KV cache. A high-bandwidth intra-node fabric (NVLink) transfers prefilled KV blocks from the prefill engine (PE) to the decode engine (DE). Our proposed design techniques can be classified into three aspects. 

First, we expose two API entrypoints: (i) \texttt{/generate}, standard prefill+decode via conventional PD pipeline following SGLang, and (ii) \texttt{/prefill\_only}, execute prefill only requests on the PE and cache the KV blocks. Notably, even we can treat prefill requests as a "zero-length-output" generate request theoretically, those requests will also introduce massive KV block transmissions from PE to DE while never being processed. This could introduce large communication latency and memory allocations, thus we separate the prefill API to avoid this situation.

\begin{figure}[t!]
    \centering
    \includegraphics[width=1.00\linewidth]{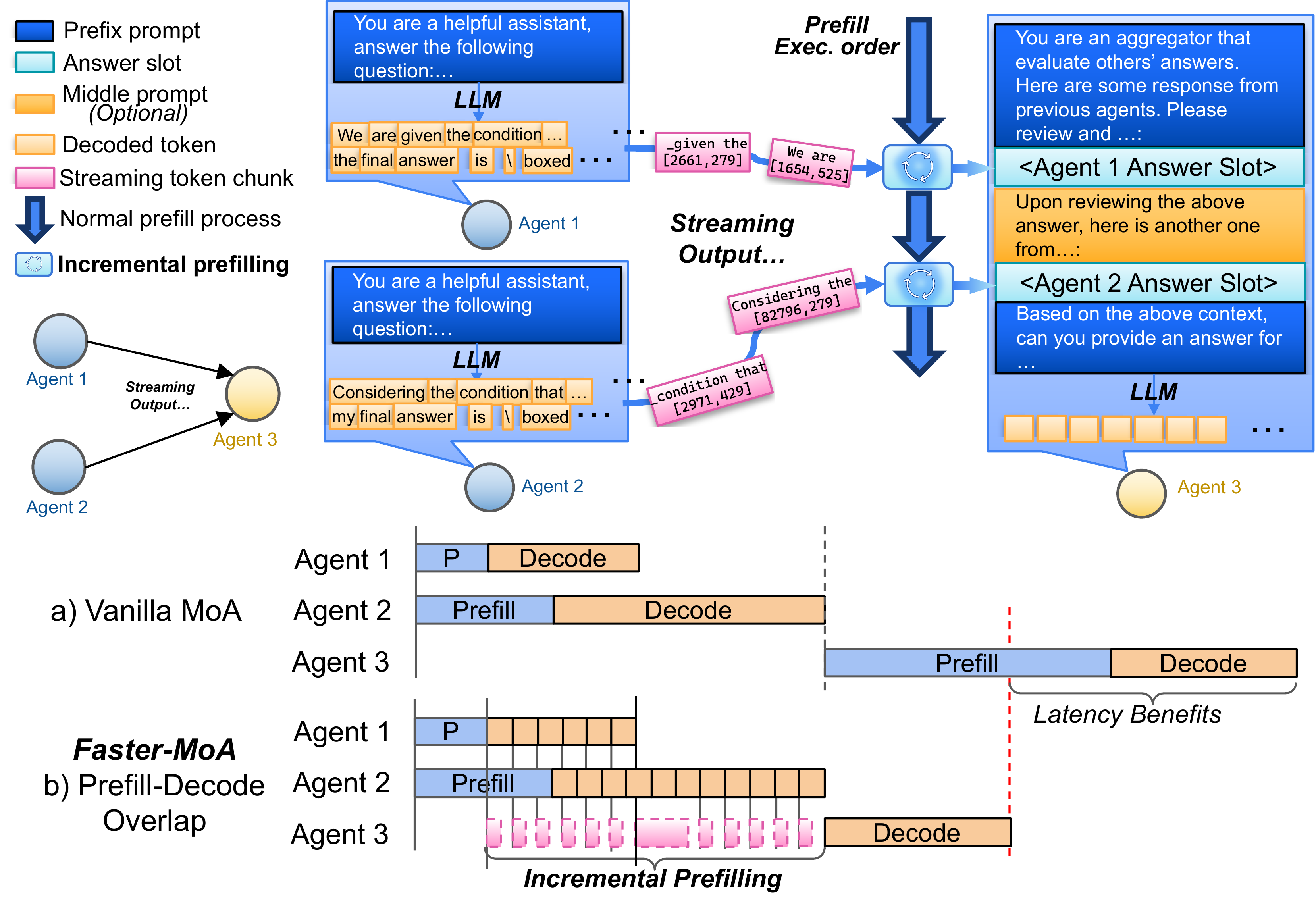}
    %\vspace{-6mm}
    \caption{Demonstration of Faster-MoA's incremental prefilling mechanism: Top: Workflow breakdown of a 3-agent dependency. Bottom: Execution bubble comparison between a) vanilla MoA, b) Our prefill-decode overlap pipeline.}
    % %\vspace{-4mm}
    \label{fig:pd-detail}
\end{figure}

Second, we develop an \emph{Agent Prompt Cache} (APC) that stores partial decoded texts for each dependent agent, enabling incremental prompt construction for successor agent prefilling. For example, in Fig.~\ref{fig:pd-detail}, partially decoded text from agent 1 and agent 2 will be stored in APC for agent 3's incremental prefilling. When using different sized models in the same model family that share the same tokenizer, APC stores the intermediate tokens instead of text, avoiding extra tokenize-detokenize process. The chunk size of partially decoded text or tokens stored in APC should be chosen empirically by users based on workload characteristics. It must balance the need to reduce successor agent’s exposed prefilling latency and the risk of generating excessive incremental prefill requests, which can incur significant prefill-engine initialization overhead. 

Third, an individual shell router outside the two engines is implemented as the core of our incremental prefilling mechanism, with details shown in Fig.~\ref{fig:1_overview}. All agents' requests will be sent to shell router first, enabling further efficient dispatching and orchestration with the native PD routers and engines. Our shell router will handle the coming agent requests in four steps: \circledwhite{1}~\textbf{Dependency identification.} If the upcoming agent request is independent, it is directly forwarded to the native PD router for direct prefilling and decoding, with decoded text/tokens streaming back to corresponding agent's APC. \circledwhite{2}~\textbf{Dependent requests handling.} If the input agent request depends on the output of other agents, the input prompt of this request will be segmented by precursor agents' output slots. The shell router then initiates incremental prefilling by first processing the prefix segment (the agent’s own prompt prefix) and continuously monitoring the first dependent agent’s APC (whose slot directly attached to the prefix) so that its corresponding segment can start prefilling immediately once the APC receives decoded tokens from that agent. \circledwhite{3}~\textbf{Incremental prefilling loop.} The shell router then periodically fetches text/token chunks from APC, concatenates them with the prefilled prefix and issues a lightweight \texttt{/prefill\_only} update. Because these increments are short and the prefix KV remains resident in HBM due to previous prefilling, incremental updates attain near 100\% KV cache hit rates. This fetch\,$\to$\,append\,$\to$\,incremental-prefill loop continues until current slot is fulfilled. If the request has multiple dependencies (as Fig.~\ref{fig:pd-detail} shown), the shell router will repeat this step until all slots are fulfilled. \circledwhite{4}~\textbf{Forward prefill-done requests.} Once all slots are filled and the agent’s input prompt is complete, its prefilling stage is already finished due to the on-the-fly overlap with precursor decoding. The router then forwards the subsequent decoding request for this agent to the internal PD router, which proceeds through the standard \texttt{/generate} path.

\section{Experimental Evaluation}
% Setup, datasets, models, serving stack, metrics (TTFT/E2E/throughput/accuracy), and results.
\subsection{Experiment Setup}
\label{sec:exp-setup}
In order to collect convincing data, we implemented our framework on two independent LLM serving engines. For accurate latency metrics, we modified SGLang v0.5.3 \cite{sglangpaper} and integrated our overlapping and early-exit mechanism, with single concurrency for exact per-sample latency. For large-batch dataset-wise verification, we adapted our framework upon vLLM v0.11.0 \cite{Kwon2023PagedAttention}, with careful prompt designs and hyperparameter tunings, and enabled concurrency=32 questions/batch to accelerate verification.

\textbf{Datasets.} We evaluated our framework on five datasets: \emph{GSM8K} \cite{cobbe2021gsm8k}, \emph{MATH-500}, \emph{AIME2025}, \emph{MMLU-ProX-Lite} and \emph{IFBench}, covering a vast majority of categories while emphasizing math reasoning and scientific Q\&A. In which, \emph{GSM8K}, \emph{MATH-500} and \emph{AIME25} represent math problems from easy to hard, \emph{MMLU-Prox-Lite} provides a wide set of general scientific questions within multiple STEM majors, and \emph{IFBench} is an instruction following testbench as an addition.

\textbf{Models.} For implementation simplicity, we chose the \emph{Qwen} model family as our candidate model pool to avoid extra heterogeneous tokenizer orchestration issues. From the pool, we finally picked three models: \emph{Qwen3-VL-4B-Instruct}, \emph{Qwen3-VL-8B-Instruct}, and \emph{Qwen3-VL-32B-Instruct}, plus an extra \emph{Qwen3-Embedding-4B} in dynamic EE routing. These three state-of-the-art models exhibit great reasoning ability in text generation \cite{qwen3technicalreport}, while their increasing dense weights perfectly show the direct proportion to growth in performance, making it a good fit for our prerequisites. On both engines, we applied the same sampling parameters as suggested in model cards for consistency.

\textbf{Hardware.} We run the models on six NVIDIA H200 GPUs (within a single NVIDIA H200 HGX Server), configured as one prefill engine and one decode engine per model. The maximum output tokens were capped at 65535, and the scheduling conservativeness was set to 0 to enforce aggressive request scheduling in native SGLang Router~\cite{sglrouterdoc}, further maximizing memory utilization.

\subsection{Dynamic EE and Incremental Prefill Impact}
% In this section, we analyze the individual impact of dynamic agent-level early exit and incremental prefilling on end-to-end latency. 
In this section, we analyze the individual impact of dynamic agent-level early exit and incremental prefilling on end-to-end latency. We first employ each mechanism in isolation under identical settings to quantify its standalone role in reducing the critical path. We then study their combined effect in Sec.~\ref{sec:full}, highlighting how early exit and incremental prefilling interact to further shrink exposed latency without accuracy degradation.

\subsubsection{Dynamic EE}
Before the integration of our dynamic early-exit method, the following questions still remained unsolved: \textbf{1) What is the explicit criterion of dynamic EE?} \textbf{2) How much latency overhead will dynamic EE introduce?}

With these in mind, we carried out experiments with only tree topology + dynamic EE deployed, and verified the framework on the five datasets. To visualize the process of dynamic EE, we collected the activation percentage of each model (4B, 8B and 32B) by calculating \texttt{invoked\_times}/\texttt{total\_samples}, as well as detailed latency breakdowns in performing dynamic EE.

\begin{figure}[t!]
    \centering
    % ---------------- Top row: left + right ----------------
    \begin{subfigure}[c]{0.49\linewidth}
        \centering
        \includegraphics[width=\linewidth]{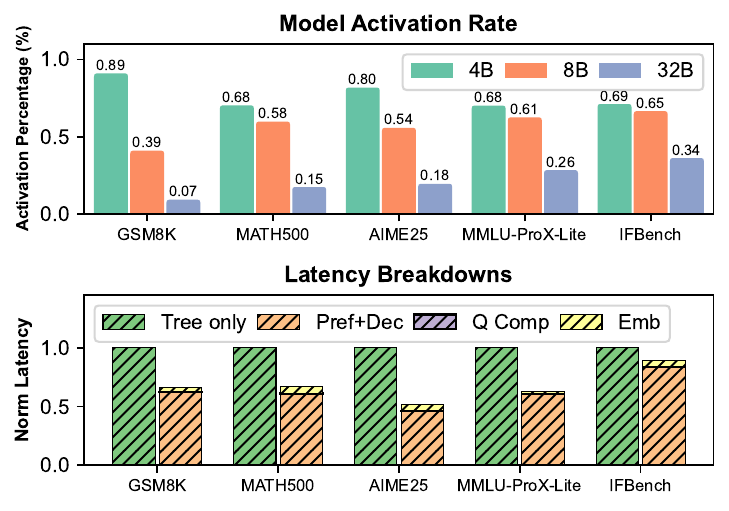}
        \caption{Activation percentage of three candidate models on datasets (top) and dynamic EE latency breakdown (bottom).}
        \label{fig:latency_bd_and_act_times}
    \end{subfigure}
    \hfill
    \begin{subfigure}[c]{0.48\linewidth}
        \centering
        \includegraphics[width=\linewidth]{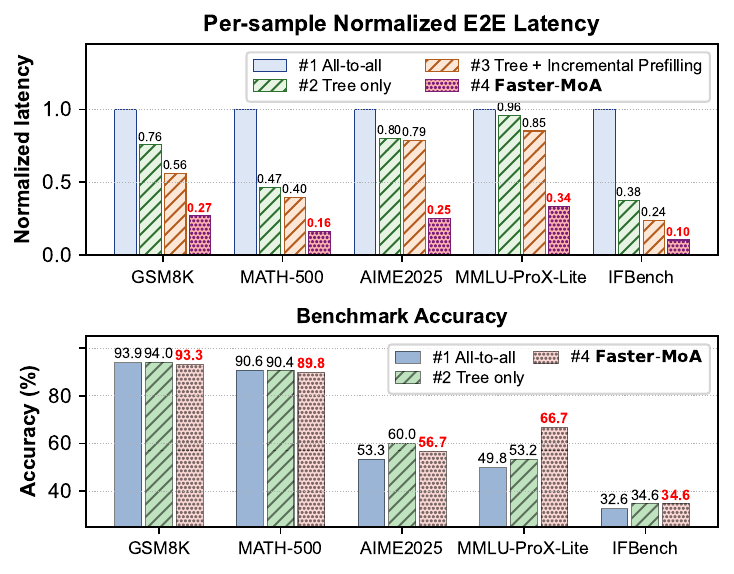}
        \caption{Experiment results of FaSTeR-MoA on datasets. Top: per-sample normalized E2E latency. Bottom: overall test accuracies.}
        \label{fig:final}
    \end{subfigure}

    \vspace{0.8em}

    % ---------------- Bottom row: single centered ----------------
    \begin{subfigure}[c]{0.6\linewidth}
        \centering
        \includegraphics[width=\linewidth]{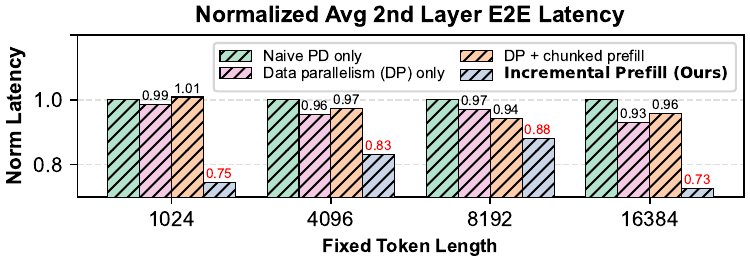}
        \caption{Normalized average second-layer E2E latency with different prefill–decode optimizations.}
        \label{fig:pd_preliminary}
    \end{subfigure}

    \caption{Summary of FaSTeR-MoA performance and ablation results across benchmarks.}
    \label{fig:all_results}
\end{figure}

As illustrated in Fig.~\ref{fig:all_results}(\subref*{fig:latency_bd_and_act_times}), we observed that: \textbf{(i)} With dynamic EE, large model is significantly less likely to be invoked compared to small models. \textbf{(ii)} For harder datasets compared with easier ones (e.g, \emph{IFBench} to \emph{GSM8K}), more invocations of large model happen in IFBench, the harder one. These two observations validate the effectiveness of score Q, which guides the early exit process adaptively towards different-level questions. Additionally, as observed from
the latency breakdowns, the calculation in our proposed method would only introduce $\sim$5\% additional latency, however brings about $10\% \sim 50\%$ E2E latency reduction in total, thus the effectiveness is ensured.

%\vspace{-2mm}
\subsubsection{Incremental Prefilling}
We then investigated the latency metrics on \textbf{second} layer aggregators in the tree structure with random tokens. This specified metric is critical in our framework, since it explicitly impacts the next final aggregator's input. If we accelerate the E2E latency for all second layer aggregators, we can establish an early decision in our dynamic EE routing.

We consider three model deployment baselines in contrast to our proposed method here: \textbf{1) Naive PD disaggregation only. } \textbf{2) Data parallelism (DP) only}. Since we are utilizing two GPUs for a single model, the worth of resources is vital. \textbf{3) DP + chunked prefill \cite{agrawal2023sarathi}}, as an addition to 2). We use random generated tokens as input with uniformly distributed length $L \sim U(1, 2048)$, and control output token length (denoted as \texttt{fixed\_tokens}), then collect the E2E latency on the second layer.

As shown in Fig.~\ref{fig:all_results}(\subref*{fig:pd_preliminary}), our proposed method generally outperformed the three baseline, with a maximum of 27.4\% E2E latency reduction while the baselines reaching only $\sim$10\%. The above concretely illustrates the feasibility of our proposed overlapping method.

\subsection{Final Experiment \& Result Discussions}
\label{sec:full}
\textbf{Settings.} For our ultimate experiment, we considered the following four settings: \textbf{1) All-to-all Baseline:} three layers with 9 agents $\rightarrow$ 9 agents $\rightarrow$ aggregator. \textbf{2) Tree structure only:} replace baseline with our three-layered 9-3-1 tree structure, without further optimizations. \textbf{3) Tree structure + Incremental Prefilling:} only enables the incremental prefilling method. \textbf{4) fully-integrated \textbf{Faster-MoA} framework.}

Figure \ref{fig:all_results}(\subref*{fig:final}) illustrates the final experiment results. We observed a significant E2E latency reduction in each category. Maximum reduction of $\sim$62\% with tree-only, $\sim$76\% with tree+overlapped prefill and $\sim$90\% with fully integrated setting is achieved by the proposed frameworks, indicating the effectiveness of our methodology.

We also benchmark accuracy to verify that these optimizations do not compromise reasoning quality (Fig.~\ref{fig:final} (bottom)). Since incremental prefilling only restructures the timing of prefill and decode without modifying any prompt content, it does not affect the model’s accuracy performance; therefore, we focus our accuracy comparison on the tree-only and the fully integrated configuration versus the all-to-all baseline.

Across the five benchmarks, our fully integrated Faster-MoA achieves accuracy comparable to the all-to-all baseline on \emph{GSM8K}, \emph{MATH-500}, and \emph{IFBench}, within an acceptable $\pm 1\%$ absolute margin. This behavior is consistent with the saturation effect of aggregating many agents on relatively easier tasks, as discussed in Fig.~\ref{fig:width_saturation}. For the remaining two datasets, \emph{MMLU-ProX-Lite} and \emph{AIME2025}, we even observe noticeably higher accuracy than the all-to-all baseline. These gains suggest that dynamic EE can selectively truncate redundant or low-quality answers produced by weaker proposer agents, yielding a more reliable aggregate answer. Assembling together, these results demonstrate that Faster-MoA can substantially reduce latency while preserving or even improving end-task accuracy.

\section{Conclusion}
In this paper, we propose \textbf{Faster-MoA}, a unified algorithm-system co-design for efficient MoA serving. Our hierarchical tree-structured agent topology replaces all-to-all connectivity, substantially reducing redundant interactions through structural sparsity while still enabling both localized and global information aggregation. A run-time dynamic agent early-exit mechanism further prunes unnecessary agent connections based on both output similarity and confidence. To further improve hardware efficiency, we introduce dependency-aware incremental prefilling, which overlaps prefilling and decoding stages during inference across dependent agents. Together, these techniques enable Faster-MoA to reduce end-to-end serving latency by up to 90\% while maintaining comparable (within $\pm$1\%) or higher task accuracy compared to all-to-all MoA baselines.

\bibliographystyle{unsrt} 
\bibliography{references_fixed}

\end{document}